\documentclass{article}
\usepackage{ijcai23}

\usepackage{times}
\usepackage{soul}
\usepackage{url}
\usepackage[hidelinks]{hyperref}
\usepackage[utf8]{inputenc}
\usepackage[small]{caption}
\usepackage{graphicx}
\usepackage{amsmath}
\usepackage{amsthm}
\usepackage{booktabs}
\usepackage{algorithm}
\usepackage{algorithmic}
\usepackage[switch]{lineno}
\usepackage{amssymb}

\newcommand\equalcontribution{\thanks{Equal contribution.}}

\title{Temporal Segment Transformer for Action Segmentation}

\author{
Zhichao Liu$^1$\equalcontribution\and
Leshan Wang$^1$\footnotemark[1]\and
Desen Zhou$^1$\footnotemark[1]\and
Jian Wang$^2$\and
Songyang Zhang$^4$\and
Yang Bai$^3$\and
Errui Ding$^2$\And
Rui Fan$^1$
\affiliations
$^1$ShanghaiTech University\\
$^2$Department of Computer Vision Technology (VIS), Baidu Inc\\
$^3$Department of Computer Science, Durham University\\
$^4$Shanghai AI Laboratory
\emails
\{liuzhch, wanglsh, zhouds, fanrui\}@shanghaitech.edu.cn,
\{wangjian33, dingerrui\}@baidu.com
}

\begin{document}

\maketitle

\begin{abstract}
    Recognizing human actions from untrimmed videos is an important task in activity understanding, and poses unique challenges in modeling long-range temporal relations. 
  Recent works adopt a predict-and-refine strategy which converts an initial prediction to action segments for global context modeling. 
  However, the generated segment representations are often noisy and exhibit inaccurate segment boundaries, over-segmentation and other problems.  To deal with these issues, we propose an attention based approach which we call \textit{temporal segment transformer}, for joint segment relation modeling and denoising.  The main idea is to denoise segment representations using attention between segment and frame representations, and also use inter-segment attention to capture temporal correlations between segments.  The refined segment representations are used to predict action labels and adjust segment boundaries, and a final action segmentation is produced based on voting from segment masks.  We show that this novel architecture achieves state-of-the-art accuracy on the popular 50Salads, GTEA and Breakfast benchmarks.  We also conduct extensive ablations to demonstrate the effectiveness of different components of our design.
\end{abstract}

\section{Introduction}

Action segmentation is the task of assigning an action label to each frame of a minutes-long untrimmed video.  Action segmentation has been studied extensively in computer vision\cite{rohrbach2012database,kuehne2016end}, and plays a crucial role in achieving a fine-grained understanding of videos of human activities.  It has a number of applications, including in robotics, industrial anomalies detection and surveillance. 

\begin{figure}[h]
  \centering
  \includegraphics[width=1\linewidth]{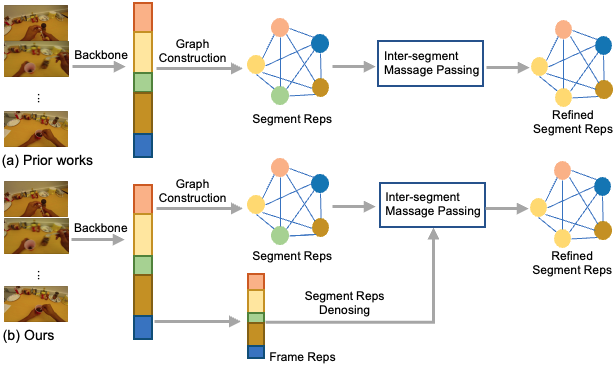}
  \caption{Comparison of different representation update strategies using prior segment modeling approaches (a) and ours (b). Prior works directly perform message passing between segments using heuristic graphs, leading to representation noise caused by inaccurate initial segment predictions. Our design jointly performs segment relation modeling and feature denoising using a unified transformer framework.}
  \label{fig:intro}
\end{figure}

Since the input to an action segmentation network usually contains thousands of frames, modeling long-range temporal relations between frames is a key challenge. 
Early works mostly followed a per-frame classification paradigm.  They first aggregate frame features in a hierarchical manner, using gradually larger temporal kernels and stacking multiple layers of per-frame representations computed using TCN\cite{farha2019ms}, GCN\cite{DTGRM} or Transformers\cite{asformer}.   However, such hierarchical frame-level modeling has several disadvantages. For example, since the method requires many layers to achieve a large receptive field, it leads to inefficient modeling of global context.  This can lead to a number of errors, including incorrect ordering of the predicted action sequence, incorrect action segment labels, over or under-segmentation and errors in segment boundaries.  Another important problem is that certain benchmark videos \cite{50salads,gtea} have a limited camera view, while others \cite{huang2020improving} show a first-person perspective from which certain actions cannot be directly observed.  In the latter case, while humans are able to infer certain unseen actions from context (for example, a ball which is shown only in the left and later right portion of the screen can be inferred to have crossed the middle), such a task is generally difficult for neural networks.

To address these problems, recent works have proposed a "predict-refine" paradigm which models relations between action segments \cite{hasr,huang2020improving}.  Specifically, they first use a backbone action segmentation network (typically based on the per-frame classification model, e.g. MS-TCN\cite{farha2019ms}) to generate an initial segmentation prediction, which is then converted into a set of temporally ordered action segments.  Next, they compute a representation for each segment, and model relationships between the segments using techniques such as graph message passing or GRU\cite{hasr}.  These relationships are used to update the segments and are mapped to corresponding frames for the final action predictions.  

Despite encouraging results, one crucial limitation in existing segment modeling approaches is that the initial segment representations are typically noisy, due to inaccurate initial frame predictions.  Hence, subsequent message passing performed on these noisy representations may propagate noise between segments and reduce accuracy. 
To overcome this problem, we propose to examine the relationship between each initial segment representation and a subsequence of other frames in order to refine both the segment predictions and frame representations.  

To implement the above intuition, we design an attention-based network which performs joint segment relations modeling and denoising.  The network uses one branch to extract frame representations and form segment representations, then denoises the segment representations using attention between each segment and other frames within a local window.  Also, in contrast to prior works \cite{huang2020improving} which use heuristic segment graphs for auxiliary tasks such as segment labeling and boundary regression, we perform these tasks using an additional inter-segment attention block.  A comparison of segment representation update strategies is shown in Figure \ref{fig:intro}.  For each refined segment representation, we next perform two tasks, namely segment classification to refine the action label of the initial prediction, and segment boundary regression to refine the boundary of the initial segment.  Lastly, we convert each segment boundary to a binary mask and fuse these with the segment classification probabilities using a voting strategy to generate the final action segmentations.

We evaluate our method on three widely used benchmarks, 50Salad\cite{50salads}, GTEA\cite{gtea} and Breakfast\cite{brakfast}. The results show that our approach outperforms prior works and achieve new state-of-the-art accuracy results.  We furthermore validate our design using a number of ablation studies which show the significant performance improvements achieved by our model's modules.  

To summarize, the contributions of this paper are as follows:
\begin{itemize}
    \item We propose to jointly model segment relations and denoise segments representations using a unified transformer architecture.
    \item The refined action segment representations are able to predict segment labels and boundaries, which are combined in a mask voting strategy.
    \item We validate the effectiveness of our design through extensive ablations, and achieve new state-of-the-art accuracy on three public benchmarks, 50Salads, GTEA and Breakfast.
\end{itemize}

\section{Related Work}

Earlier approaches to action segmentation used a sliding window to detect action segments\cite{rohrbach2012database,karaman2014fast}. 
Some approaches modeled the temporal action sequence using hidden Markov models \cite{kuehne2016end,tang2012learning}. 
In recent years temporal convolution achieved great success in speech synthesis, and motivated by this, the per-frame classification paradigm applies temporal convolutional networks (TCN) to action segmentation\cite{lea2017temporal,lea2016temporal,lei2018temporal,farha2019ms}. TCN consists of 1D dilated convolution with multiple dilation rates, giving algorithms such as MS-TCN large receptive fields and capacity and benefiting their temporal modeling capabilities. C2F-TCN\cite{singhania2021coarse} implicitly ensembles multiple temporal resolutions, which produces smoother segmentations and obviates additional refinement modules.

In contrast to the previous approaches, the "predict-refine" \cite{chen2020action,wang2020boundary,huang2020improving,ishikawa2021alleviating,hasr,li2022bridge} paradigm first uses an arbitrary segmentation backbone to produce a preliminary segmentation, then seeks to improve the accuracy of the initial output using a variety of mechanisms.  For example, HASR\cite{hasr} models the relationships between the segment-level features and the entire video context embedding to allow modifications to the labels of action segments.  However, this method does not refine the boundaries of segments, which is actually the most common source of segmentation error in many cases.  \cite{huang2020improving} constructs two parallel GCNs to refine the segment class and segment boundary respectively. 
Similarly, ASRF\cite{ishikawa2021alleviating} proposes the two parallel action segmentation and boundary regression branches and leverages the boundaries to refine the frame-wise classification, which reduces the over-segmentation problem.


Recently, transformers have also been applied in action segmentation to improve performance.  ASFormer\cite{asformer} proposed an effective hierarchical attention mechanism which is used to capture dependencies in minutes-long video sequences, and designed a decoder to refine their output.  In weakly supervised action segmentation, \cite{ridley2022transformers} uses sliding windows on frame-level features in a Transformer encoder to capture local context rather than that from the entire video sequence.



\begin{figure*}[h]
  \centering
  \includegraphics[width=\linewidth]{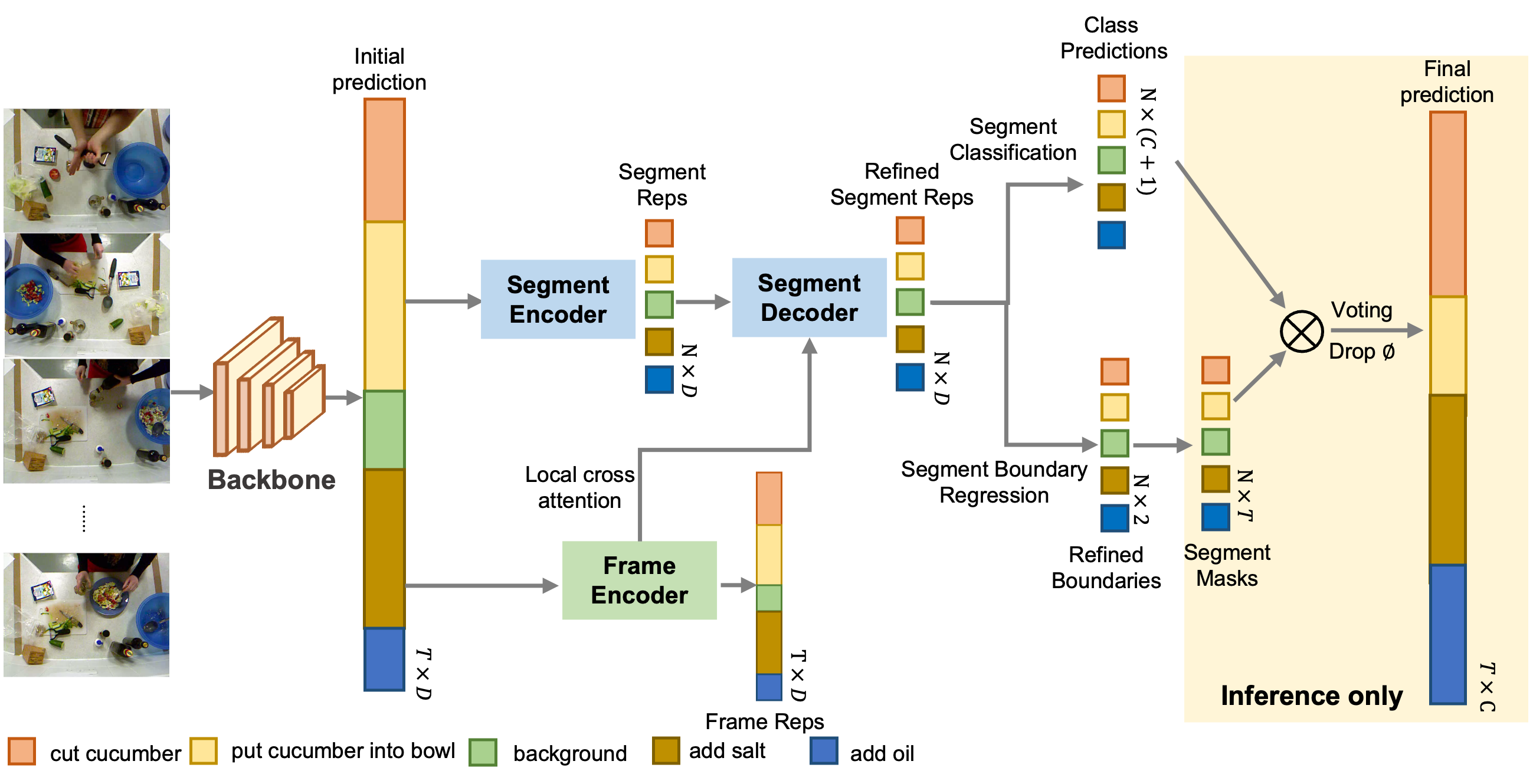}
  \caption{Overview of our framework. Our model jointly performs segment relation modeling and feature denoising in a Transformer-based architecture. The \textit{backbone network} first generates an initial segmentation prediction and frame features. The \textit{segment encoder} then generates initial segment representations. The \textit{frame encoder} further improves the frame representations for segment feature denoising. The \textit{segment decoder} exploits segment-frame and inter-segment attention for feature denoising. Each refined segment representation predicts an action probability distribution and refines segment boundaries, which are transformed to a binary segment mask. The final prediction is obtained by the voting of different segment masks.}
  \label{overview}
\end{figure*}

\section{Temporal Segment Transformer}

\subsection{Overview}

Suppose the video being segmented consist of $T$ frames, each with size $H \times W$ in RGB color; denote the video as $\mathbf{V}=\{I_t\in\mathbb{R}^{H\times W\times 3} \,  | \, t=1,2,...,T\}$.  The goal of action segmentation is to predict action labels $\mathbf{S} = \{c_t\in \mathcal{C} \,  |\, t=1,2,...,T\}$, where $\mathcal{C}$ is the set of action labels for the dataset, and $c_t$ is the action of the $t$'th frame. 
Conventional action segmentation methods often adopt a per-frame classification paradigm, which first learn a representation for each frame and then perform a $|\mathcal{C}|$-class classification on the representation, resulting in a probability distribution $p_t \in \Delta^{|\mathcal{C}|}$, where $\Delta^{|\mathcal{C}|}$ is the $|\mathcal{C}|$-dimensional probability simplex. 
Since the videos usually contain thousands of frames, modeling long-range temporal relations between frames is challenging using frame representations.
Therefore, we follow recent approaches\cite{hasr,huang2020improving} which first groups frames into segments, then models temporal relations between the action segments.  As an example of such a relation, consider the GTEA dataset containing cooking related videos, in which the ``chopping vegetables" action is often, though not always preceded by the ``washing vegetables" action.  

In more detail, we first utilize an off-the-shelf backbone, such as the state-of-the-art hierarchical model ASFormer\cite{asformer} to generate an initial segmentation prediction.  Our method is agnostic to the backbone, so that other models such as MS-TCN\cite{farha2019ms} can be used as well.  
The backbone network predicts a sequence of frame-level probability distributions $Y=[p_1, \ldots, p_T]$, and also generates frame representations $\Gamma = [\gamma_1, \ldots, \gamma_T]$, where each $\gamma_t \in \mathbb{R}^D$ is the features of the $t$'th frame and $D$ is the dimensionality of the features.  However, there are often errors in the initial backbone-produced segmentation, which we want to correct.  

Our algorithm for improving the segmentation is shown schematically in Figure \ref{overview}.  We first predict an action type for each frame $t$ using the initial frame probabilities $[p_1, \ldots, p_T]$, and then group consecutive actions of the same type together into an \emph{action segment}.  Next we form representations of each action segment using a \emph{segment encoder}, and also refine the frame representations using a computationally efficient method called the \emph{frame encoder}.  We combine the outputs of the segment and frame encoders using the \emph{segment decoder}, and then model temporal relationships between the segments using an attention mechanism between segments and local frames, as well as between pairs of segments.  This produces corrected class predictions for each action segment and also refines the boundaries of the segments.  While it is possible to first generate corrected class predictions and then use these as input to boundary refinement, we found experimentally that this only minimally improves accuracy, and thus we adopt a simpler approach of performing the tasks concurrently.  Finally, we convert the refined boundaries to segment masks delimiting the temporal extent of each segment, and then combine this with the corrected class predictions to produce our final action segmentation.

\begin{figure}[t]
  \centering
  \includegraphics[width=0.8\linewidth]{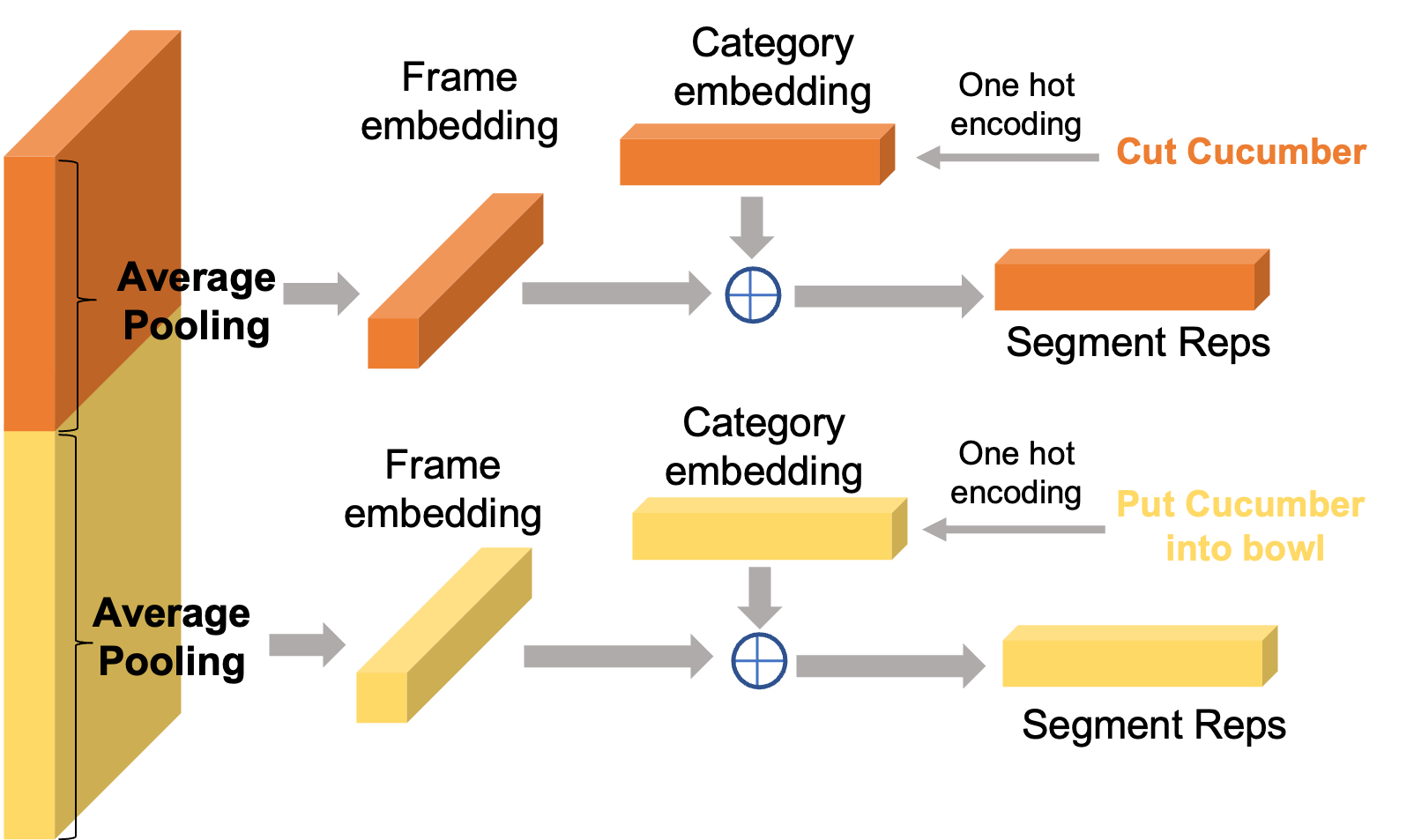}
  \caption{The segment encoder extracts segment representations based on the initial prediction and backbone features. Each representation consists of a frame embedding and a category embedding based on the initial segments.}
  \label{fig:query generation}
\end{figure}

\subsection{Segment Encoder}

The segment encoder generates segment representations $\{{\xi}_i\in\mathbb{R}^D\}_{i=1}^N$ ($N$ is the total number of segments) based on the backbone predictions $Y$ and corresponding features $\Gamma$.  We also take the class with the highest probability for each frame in $Y$ to produce a sequence of predictions $[c_1, \ldots, c_T]$, where each $c \in \mathcal{C}$.  For each segment $i \in 1, \ldots, N$, let $l_i^s$ and $l_i^e$ denote the start and end frame of the segment.  We produce a latent representation of the $i$'th segment using two terms, a frame embedding and a category embedding.  For the frame embedding, we extract all the frame representations from the $i$'th segment and average them: 
\begin{equation}
    {\xi}_{f_i} = \frac{1}{l_i^e-l_i^s+1}\sum\limits_{t=l_i^s}^{l_i^e} \gamma_t.
\end{equation}

For the category embedding, we take a one-hot vector $e_{c_i}\in \{0,1\}^{|\mathcal{C}|}$ for category $c_i$, which has the $c_i$'th coordinate set to 1, and use a multilayer perception (MLP) to map it to a latent space to produce a category representation:
\begin{equation}
    {\xi}_{c_i} = \operatorname{MLP}(e_{c_i})
\end{equation}
Finally, the frame embedding and category embedding are added to produce the segment representation:
\begin{equation}
    \mathbf{{\xi}_i} = {\xi}_{f_i} + {\xi}_{c_i}
    \label{f:seg_encoder}
\end{equation}
An illustration of the segment encoder is shown in Figure \ref{fig:query generation}.  

\subsection{Frame Encoder}

The frame encoder is another network branch which aims to improve the frame representations produced by the backbone in a computationally efficient manner.  In particular, when training our segmentation network, we optimize the frame encoder using backpropagation, but not the computationally more expensive backbone network, thereby reducing overall computational load.  We use the popular MS-TCN \cite{farha2019ms} as our frame encoder due to its simple but effective structure.  The MS-TCN network contains $L$ layers, for a hyperparameter $L$, and the receptive field of the $i$'th layer is set to $2^i$ frames to allow the network to gradually capture global context.  The frame encoder takes as input the initial frame representations $\Gamma$ and outputs 
\begin{equation}
    \hat{\Gamma} =   [\hat{\gamma}_1, \ldots, \hat{\gamma}_T] = \operatorname{MS-TCN}(\Gamma) 
    \label{f:frame_encoder}
\end{equation}

\subsection{Segment Decoder}

\paragraph{\textbf{Segment-frame Attention Block}} 
The predictions and representations produced by the backbone are sometimes noisy and inaccurate, and using only this information for segmentation may lead to poor results.  To deal with this, we perform segment representation denoising in our segment decoder using an attention mechanism between each segment and nearby frame representations. We call this form of attention \emph{segment-frame} attention.  We found it is challenging and in fact sometimes harmful for segments to attend to very distant frames.  Thus for each segment we restrict attention between the segment to only frames from its own or nearby segments, e.g. the previous and next segment.  As we described in the introduction using the example with the ``place cucumber" and ``cut cheese" actions, the features of segments and nearby frames may be correlated (or anti-correlated), and thus segment-frame attention provides signals which may help improve segment labeling accuracy.   

To implement attention, we follow \cite{vaswani2017attention} and use positional encodings with different frequencies to create a tensor $\operatorname{PE}^1 \in \operatorname{R}^{T\times D}$ encoding the position of each frame and a tensor  $\operatorname{PE}^2 \in \operatorname{R}^{N\times D}$ to encode the position of each segment. Formally, given segment representations $[\mathbf{\xi}_1, \ldots, \mathbf{\xi}_N]$, we first compute the start and end frames of all the segments  $[(l_1^s, l_1^e), \ldots, (l_N^s, l_N^e)]$, then generate an attention mask $\mathcal{M}_i\in \mathbb{R}^{1\times T}$ for $i$'th segment as

\begin{equation}
  \mathcal{M}_i(t) = 
  \begin{cases}
  0  & ~~\operatorname{if}~ l_{i-1}^s \leq t \leq l_{i+1}^e,\\
  -\infty & ~~\operatorname{otherwise} \\
  \end{cases}
  \label{local_mask}
\end{equation}

Then, the $i$'th denoised segment representation ${\xi}'_i\in \mathbb{R}^{1\times D}$ is computed using attention by

\begin{equation}
  {\xi}'_i = {\xi}_i + \operatorname{softmax}(\mathcal{M}_i + W_q{\xi}_i \mathbf{K}) \mathbf{V}^\top
  \label{seg_fra_atten}
\end{equation}

Here $W_q\in\mathbb{R}$ is a weight for the query embedding, and $\mathbf{K}, \mathbf{V}\in\mathbb{R}^{D\times T}$ are the key and value embeddings, which are computed using two auxiliary MLPs $MLP_1$ and $MLP_2$, respectively.
$$\mathbf{K} = MLP_1(\hat{\Gamma}+\operatorname{PE^1})), \; \mathbf{V} = MLP_2(\hat{\Gamma}+\operatorname{PE^1}))$$

\begin{figure}[t]
  \centering
  \includegraphics[width=\linewidth]{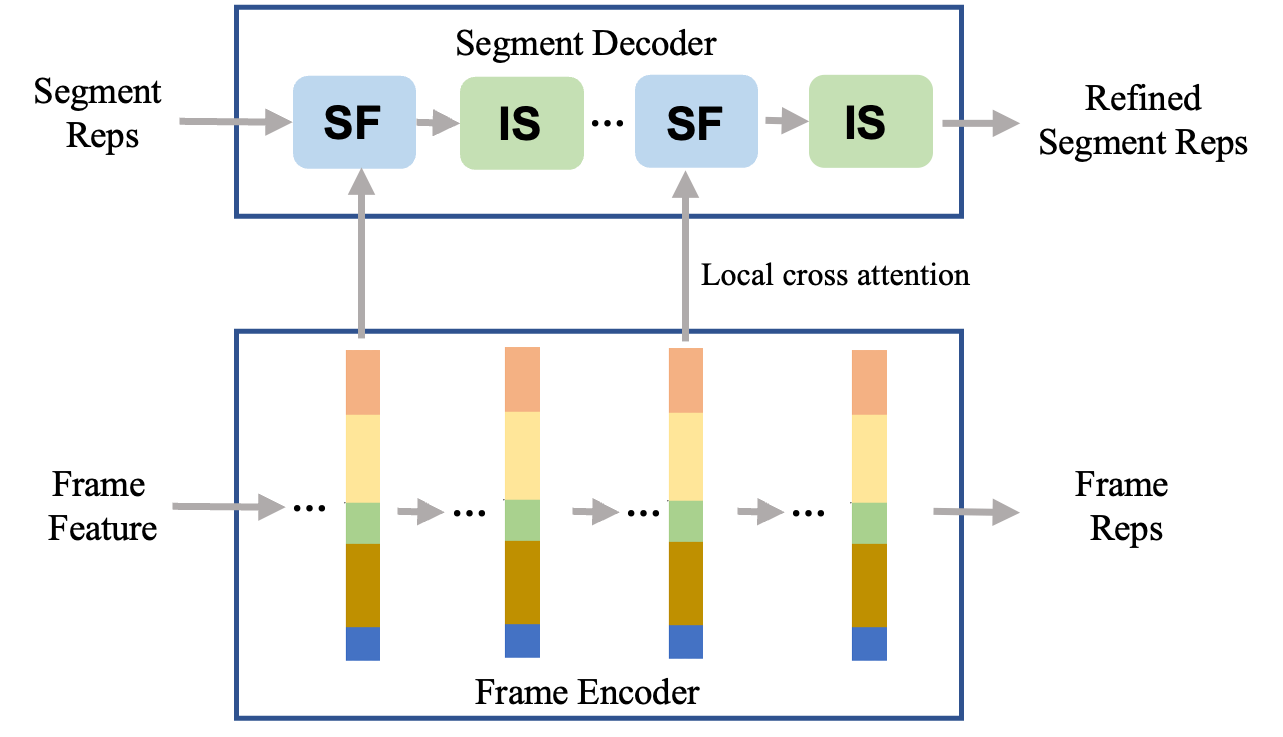}
  \caption{The segment decoder alternatingly performs segment-frame attention (SF) for segment relation modeling and inter-segment attention (IS) block for segment representation denoising.}
  \label{fig:decoder structure}
\end{figure}

\paragraph{\textbf{Inter-segment Attention Block}}
In addition to attention between segments and nearby frames, another informative type of attention is between different segments.  We call this form of attention \emph{inter-segment attention}.  Indeed, as we mentioned earlier using the example with washing and cutting vegetables, action segments often exhibit a degree of temporal (anti-)correlation, so that inter-segment attention helps constrain plausible action sequences and improves prediction accuracy.  

To implement inter-segment attention, prior works either used a one-layer GRU\cite{hasr} or heuristic segment graphs \cite{huang2020improving} for inter-segment message passing.  In contrast, we use a Transformer-based attention mechanism. 

\begin{equation}
  {\xi}''_i = {\xi}'_i + \operatorname{softmax}({W_q}'({\xi}'_i + \operatorname{PE^2}_i) )(\mathbf{{K'})^\top}) \mathbf{V'}
  \label{seg_seg_atten}
\end{equation}

Here, ${\xi}''_i \in \mathbb{R}^{1\times D}$ is the category denoised segment representation.
${W_q}'\in\mathbb{R}$ is a weight for the query embedding, $\operatorname{PE}^2 \in\operatorname{R}^{N\times D}$ is a positional encoding used for inter-segment attention, and $\mathbf{K'}, \mathbf{V'}\in\mathbb{R}^{N\times D}$ are the key and value embeddings, computed using two auxiliary MLPs $MLP_3$ and $MLP_4$ respectively.

$$\mathbf{K'} = MLP_3({\xi}'_i+\operatorname{PE^2}_i), \; \mathbf{V'} = MLP_4({\xi}'_i+\operatorname{PE^2}_i)$$

As illustrated in Figure \ref{fig:decoder structure}, the segment decoder has multiple layers and alternatingly performs inter-segment attention for segment relation modeling and segment-frame attention for segment representation denoising. 

We denote the final collection of $N$ refined segment representations produce by the segment decoder output as $\Xi'' \in\mathbb{R}^{N\times D}$.

\subsection{Prediction heads}

\paragraph{\textbf{Segment classification}} 
Having computed refined segment representations using the above procedures, we now generate refined probability distributions for the segment labels.  We use a two-layer MLP followed by a $\operatorname{softmax}$ operation.
\begin{equation}
    \mathcal{P}^{mask} = \operatorname{softmax}\left(MLP(\Xi'')\right)
    \label{segment_cls}
\end{equation}
Here $\mathcal{P}^{mask}$ is an  $N \times (|\mathcal{C}|+1)$ matrix giving the classification probabilities for each of the $N$ segments among the $C$ classes, as well as an additional ``junk" class for segments which cannot be classified.  

\paragraph{\textbf{Segment Boundary Regression}}
In addition to correcting segment classifications, we also want to correct segment boundaries using boundary regression.  For this, we follow previous works \cite{huang2020improving,gao2017cascaded} and apply a two-layer MLP on the refined segment representations.
\begin{equation}
    \hat{\mathcal{O}} =MLP(\Xi'')
    \label{mask_mapping}
\end{equation}
Here $\hat{\mathcal{O}}\in{\mathbb{R}}^{N \times 2}$ gives the boundary offsets of all segments. In particular, each segment has an offset vector $\hat{\boldsymbol{o}_i}=\left(\hat{o}_{i, c}, \hat{o}_{i, l}\right)$, where $\hat{o}_{i, c}$ is the offset of the segment center (normalized by the length of the segment), and $\hat{o}_{i, l}$ is the change in the segment's length given in log scale. 

During inference, we can use the refined boundaries $[(\hat{l}_1^s, \hat{l}_1^e), \ldots, (\hat{l}_N^s, \hat{l}_N^e)]$ to generate refined masks for the segments $\mathcal{M}' \in \mathbb{R}^{N\times T}$, where each binary mask is 1 in the frames the corresponding segment occupies and 0 otherwise.

\paragraph{\textbf{Mask Voting for Action Segmentation}}
Our ultimate goal is to obtain the frame level probability distribution over the actions 
$\mathcal{P} = \{c_t\in \mathcal{C} \,  |\, t=1,2,...,T\}$, where $\mathcal{C}$ is the set of action labels for the dataset, and $c_t$ is the action of the $t$'th frame.  We observed experimentally that directly maximizing per segment class prediction likelihood and mapping refined boundaries leads to poor performance.
Since we calculated the segment masks $\mathcal{M}'$ and their label probabilities $\mathcal{P}^{mask}$, we use a voting strategy to fuse different masks. Specifically, we drop the ``junk" class from $\mathcal{P}^{mask}$ to form matrix $\tilde{\mathcal{P}}^{mask} \in \mathbb{R}^{N \times |\mathcal{C}|}$, and then for every frame $t$ and class $c$, we calculate the frame level action probability $\mathcal{P}(t, c)$ using a weighted summation over different queries
\begin{equation}
\mathcal{P}(t, c) = \sum\limits_{i=1}^{N} \mathcal{M}'(i,t)\cdot \tilde{\mathcal{P}}^{mask}(i,c)
\end{equation}
Writing this in matrix form, we have 
\begin{equation}
    \mathcal{P}^{frame} = {(\mathcal{M}')^\top \times \tilde{\mathcal{P}}}^{mask}
    \label{eq:mask voting}
\end{equation}
Note that we do not need to apply the normalization in Eq. \ref{eq:mask voting}, since the final labels can be directly read out by taking the category with the largest value for each segment.

\subsection{Model Training}

In the segment encoder stage, for each segment representation, we first assign the ground truth segment(s). We compute the temporal intersection-over-union (tIoU) between the initial predicted segment masks and all ground truth segments. Then we perform a Hungarian matching based on the tIoU distance to best align predicted and ground truth segments. Additionally, we drop matching pairs where the matched tIoU is 0. We consider an initial segment positive if it is matched to a ground truth segment. 

To optimize the temporal segment transformer, we use losses on both segment classification and boundary regression accuracy.  We use cross entropy loss $L_{ce}$ for the classification task. Our regression strategy follows \cite{huang2020improving} and uses smooth L1 loss in the loss term $L_{reg}$.  To generate ground truth offsets for the boundary regression head, we let $l_{i, c}=\left(l_{i} ^s+l_{i}^e\right) / 2$ be the center of segment $i$ and let $l_{i, l}=l_{i}^e-l_{i}^s$ be its length.  The ground truth offset $\boldsymbol{O}_{i}^{g t}=\left(o_{i, c}^{g t}, o_{i, l}^{g t}\right)$ is
\begin{equation}
    o_{i, c}^{g t}= \frac{l_{i, c}-l_{i, c}^{g t}}{l_{i, l}}, \quad o_{i, l}^{g t}=\log  \left(\frac{l_{i, l}}{l_{i, l}^{g t}}\right)
    \label{eq:loss}
\end{equation}

The overall decoder loss function has the form:
\begin{equation}
    \label{eq-loss}
    \mathcal{L} = \lambda_{1}L_{ce} + \lambda_{2} L_{reg}(\mathcal{O}^{g t}, \hat{\mathcal{O}}), 
\end{equation}
where $\lambda_{1}, \lambda_{2}$ are hyperparameters.

\section{Experiments}


\subsection{Datasets}
\textbf{50Salads\cite{50salads}} 
This dataset consists of 50 videos with 17 action classes and contains 20 action instances which were performed by 25 human subjects, and videos are 6.4 minutes long on average. The 5-fold cross-validation is performed for evaluation.

\textbf{GTEA\cite{gtea}}
This dataset consists of 28 egocentric videos with 11 action classes of daily activities in the kitchen performed by 4 human subjects.
Each video has an average of 20 action instances, with an average duration of about half a minute. The 4-fold cross-validation is performed for evaluation

\textbf{Breakfast\cite{brakfast}} This dataset consists of 1712 videos of 18 different activities in kitchens, showing breakfast preparation from 52 human subjects.  The dataset can be divided into four splits. The videos are annotated with 48 different actions and contain 6 action instances on average. The 4-fold cross-validation is performed for evaluation.


\subsection{Evaluation Metrics}
We evaluate performance using several metrics, including frame-level accuracy (Acc), a segmental edit score (Edit) and segmental F1 score with overlap threshold $k/100$, for $k=10, 25, 50$. 
The edit score penalizes over-segmentation, while the segmental F1 score measures prediction quality. 
For each dataset,  we use k-fold cross validation and report average results.

\subsection{Implementation Details}
We freeze the backbone and train our Transformer head for 60 epochs. 
We used a similar training strategy as prior work \cite{hasr} which utilized the data splits and predictions from early stop epochs. 
The learning rate was set to 1e-4 on all datasets. 
All experiments used the Adam optimizer with weight decay rate 1e-4.

We use MS-TCN\cite{farha2019ms} as our frame encoder with $L = 10$ layers and with the size of the local window doubled at each layer, starting from size 2.  We used $D=64$ as the dimension of hidden representations.  For the segment decoder, we increased the backbone feature dimension to $D=256$, and used two layers which corresponded to frame features from the 8'th and 9'th layers of the frame encoder.
The weights $\lambda_1$ and $\lambda_2$ for the different loss components in Eq.\ref{eq-loss} were set to 1.

\begin{table}[t]
	\centering
	\caption{Performance comparison on 50Salads dataset.}
	\resizebox{0.5\textwidth}{!}{
	\begin{tabular}{cccccc}
		\hline
		\textbf{50Salads} &\multicolumn{3}{c}{\textbf{F1}$@\{ \text{10,25,}50 \}$} &\textbf{Edit} &\textbf{Acc} \\  
		\hline
		IDT+LM~\cite{IDTLM} &44.4 &38.9 &27.8 &45.8 & 48.7 \\
		ST-CNN~\cite{STACNN} &55.9 &49.6 &37.1 &45.9 & 59.4 \\
		Bi-LSTM~\cite{bilstm} &62.6 &58.3 &47.0 &55.6 & 55.7 \\
		ED-TCN~\cite{lea2017temporal} &68.0 &63.9 &52.6 &59.8 & 64.7 \\
		TDRN~\cite{lei2018temporal} &72.9 &68.5 &57.2 &66.0 & 68.1 \\
		SSA-GAN~\cite{SSAGAN} &74.9 &71.7 &67.0 &69.8 & 73.3 \\
		MS-TCN~\cite{farha2019ms} &76.3 &74.0 &64.5 &67.9 & 80.7 \\
    MS-TCN~\cite{farha2019ms}(HSAR impl) &77.2 &74.7 &64.8 &70.4 & 80.3 \\
    MS-TCN~\cite{farha2019ms} + HASR\cite{hasr} &83.4 &81.8 &71.9 &77.4 & 81.7 \\
		DTGRM~\cite{DTGRM} &79.1 &75.9 &66.1 &72.0 &80.0 \\
		BCN~\cite{wang2020boundary} &82.3 &81.3 &74.0 &74.3 & 84.4 \\
		Gao \cite{GAO} &80.3 &78.0 &69.8 &73.4 & 82.2 \\
		ASRF~\cite{ishikawa2021alleviating} &84.9 &83.5 &77.3 &79.3 & 84.5 \\
		HASR\cite{hasr}+ASRF\cite{ishikawa2021alleviating} & 86.6 & 85.7 & 78.5 & 81.0 & 83.9 \\
		ASFormer~\cite{asformer} &85.1 &83.4 &76.0 &79.6 & 85.6 \\
		ASFormer~\cite{asformer} + HASR\cite{hasr}(our impl.)&86.2 & 84.5 & 77.2 &81.3 & 85.3 \\
		ASFormer~\cite{asformer} + ASRF\cite{ishikawa2021alleviating} & 86.8 &85.4 &79.3 &81.9 & 85.9 \\
		\hline
    MS-TCN~\cite{farha2019ms} +   Ours &\textbf{83.9} &\textbf{82.7} &\textbf{72.9} &\textbf{78.5} &\textbf{82.6} \\
		ASFormer~\cite{asformer} + Ours &\textbf{87.1} &\textbf{85.9} &\textbf{78.3} &\textbf{82.4} &\textbf{86.1} \\
		ASFormer~\cite{asformer} + ASRF\cite{ishikawa2021alleviating} + Ours &\textbf{87.9} &\textbf{86.6} &\textbf{80.5} &\textbf{82.7} &\textbf{86.6} \\
		\hline
	\end{tabular}}
	\label{50salads_sota}
\end{table}

\begin{table}[t]
	\centering
	\caption{Performance comparison on GTEA dataset.}
	\resizebox{0.5\textwidth}{!}{
	\begin{tabular}{cccccc}
		\hline
		\textbf{GTEA} &\multicolumn{3}{c}{\textbf{F1}$@\{ \text{10,25,}50 \}$} &\textbf{Edit} &\textbf{Acc} \\  
		\hline
		ST-CNN~\cite{STACNN} &58.7 &54.5 &41.9 &- & 60.6 \\
		Bi-LSTM~\cite{bilstm} &66.5 &59.0 &43.6 &- & 55.5 \\
		ED-TCN~\cite{lea2017temporal} &72.2 &69.3 &56.0 &- & 64.0 \\
		TDRN~\cite{lei2018temporal} &79.2 &74.4 &62.7 &74.1 & 70.1 \\
		SSA-GAN~\cite{SSAGAN} &80.6 &79.1 &74.2 &76.0 & 74.4 \\
		MS-TCN~\cite{farha2019ms} &85.8 &83.4 &69.8 &79.0 & 76.3 \\
    MS-TCN~\cite{farha2019ms}(hasr impl) &88.6 &86.4 &72.5 &83.9 & 78.3 \\
    MS-TCN~\cite{farha2019ms} + HASR\cite{hasr} &89.2 &87.3 &73.2 &85.4 & 77.4 \\
		DTGRM~\cite{DTGRM} &87.8 &86.6 &72.9 &83.0 &77.6 \\
		BCN~\cite{wang2020boundary} &88.5 &87.1 &77.3 &84.4 & 79.8 \\
		Gao \cite{GAO} &89.9 &87.8 &75.8 &84.6 & 78.5 \\
		ASRF~\cite{ishikawa2021alleviating} &89.4 &87.8 &79.8 &83.7 & 77.3 \\
		HASR\cite{hasr} + ASRF\cite{ishikawa2021alleviating} & 89.2 & 87.2 & 74.8 & 84.5 & 76.9 \\
		ASFormer~\cite{asformer} &90.1 &88.8 &79.2 &84.6 & 79.7 \\
		ASFormer~\cite{asformer} + HASR\cite{hasr}(our impl.)&90.7 &89.4 & 80.3 & 85.8 & 79.3 \\
		\hline
    MS-TCN~\cite{farha2019ms} + Ours &\textbf{90.1} &\textbf{87.9} &\textbf{74.4} &\textbf{86.1} &\textbf{78.4} \\
		ASFormer~\cite{asformer} + Ours &\textbf{91.4} &\textbf{90.2} &\textbf{82.1} &\textbf{86.6} &\textbf{80.3} \\
		\hline
	\end{tabular}}
	\label{gtea_sota}
\end{table}

\subsection{Comparisons with the State-of-the-Art Methods}

We compare our temporal segment transformer with a number of existing state of the art action segmentation methods in Tables \ref{50salads_sota}, \ref{gtea_sota} and \ref{breakfast_sota}. The results show that our method outperforms all prior methods on all datasets.

The results on the \textbf{50Salads} dataset are shown in Table \ref{50salads_sota}. Our model exceeds the prior state of the art model ASFormer\cite{asformer} with HASR\cite{hasr} by $0.9$\%, $1.4$\% and $1.1$\% on segmental F1 score with different $k$ values, by $1.1$\% on segment edit distance, and by $0.8\%$ on frame-wise accuracy. 
Since ASFormer is improved by incorporating the boundary refinement algorithm ASRF\cite{ishikawa2021alleviating}, 
we also experimented with combining our model and ASRF. 
The last row in Table \ref{50salads_sota} shows that adding ASRF to our method further improves segmental F1 score by $0.8\%$, $0.7\%$ and $1.9\%$, improves segment edit distance by $1.3\%$,  and frame-wise accuracy by  $0.5\%$. 
We also experimented with using MS-TCN~\cite{farha2019ms} as a backbone.  As shown in the third to last line in Table \ref{50salads_sota}, our method outperforms MS-TCN with HASR by $0.5$\%, $0.9$\% and $1.0$\% on segmental F1 score, $1.1$\% on segment edit distance, and $0.9\%$ on frame-wise accuracy.

On the \textbf{GTEA} dataset, our model outperforms ASFormer with HASR by  $0.7$\%, $0.8$\% and $1.8$\% on segmental F1 score, by $0.8$\% on segment edit distance, and by $1.0\%$ on frame-wise accuracy. Since F1 score and edit distance measure the quality of the segments, the significant improvements we obtain demonstrate the effectiveness of our mask classification strategy.
Our model also outperform MS-TCN with HASR by $0.9$\%, $0.6$\% and $1.2$\% on segmental F1 score, by $0.7$\% on segment edit distance, and $1.0\%$ on frame-wise accuracy.

On the \textbf{Breakfast} dataset, our method  exceeds ASFormer with HASR by $1.2$\%, $1.1$\% and $1.0$\% on segmental F1 score, by $1.2$\% on segment edit distance, and by $1.5\%$ on frame-wise accuracy.
Using MS-TCN as a backbone, we outperform MS-TCN with HASR $1.1$\%, $0.9$\%, $1.3$\% on segmental F1 score, by $0.8$\% on segment edit distance, and by $1.2\%$ on frame-wise accuracy.

\begin{table}[t]
  \centering
	\caption{Performance comparison on Breakfast dataset.}
	\resizebox{0.5\textwidth}{!}{
	\begin{tabular}{cccccc}
		\hline
		\textbf{Breakfast} &\multicolumn{3}{c}{\textbf{F1}$@\{ \text{10,25,}50 \}$} &\textbf{Edit} &\textbf{Acc} \\  
		\hline
		ED-TCN~\cite{lea2017temporal} &- &- &- &- & 43.3 \\
        HTK~\cite{HTK} &- &- &- &- & 50.7 \\
        TCFPN~\cite{TCFPN} &- &- &- &- & 52.0 \\
        SA-TCN~\cite{SATCN} &- &- &- &- & 50.0 \\
        HTK(64) ~\cite{kuehne2016end} &- &- &- &- & 56.3 \\
		MS-TCN~\cite{farha2019ms} &52.6 &48.1 &37.9 &61.7 & 66.3 \\
    MS-TCN~\cite{farha2019ms}(HASR impl) &63.5 &58.3 &45.9 &66.2 & 67.7 \\
    MS-TCN~\cite{farha2019ms} + HASR\cite{hasr} &73.2 &67.9 &54.4 &70.8 & 69.8 \\
		DTGRM~\cite{DTGRM} &68.7 &61.9 &46.6 &68.9 &68.3 \\
		BCN~\cite{wang2020boundary} &68.7 &65.5 &55.0 &66.2 & 70.4 \\
		Gao \cite{GAO} &74.9 &69.0 &55.2 &73.3 & 70.7 \\
		ASRF~\cite{ishikawa2021alleviating} &74.3 &68.9 &56.1 &72.4 & 67.6 \\
		HASR\cite{hasr} + ASRF\cite{ishikawa2021alleviating} & 74.7 & 69.5 & 57.0 & 71.9 & 69.4 \\
		ASFormer~\cite{asformer} &76.0 &70.6 &57.4 &75.0 & 73.5 \\
		ASFormer~\cite{asformer} + HASR\cite{hasr}(our impl.)&76.3 &71.2 & 58.5 &74.5 & 72.2 \\
		\hline
    MS-TCN~\cite{farha2019ms} + Ours &\textbf{74.3} &\textbf{68.8} &\textbf{55.7} &\textbf{71.6} &\textbf{71.0} \\
		ASFormer~\cite{asformer} + Ours &\textbf{77.5} &\textbf{72.3} &\textbf{59.5} &\textbf{76.7} &\textbf{73.7} \\
		\hline
	\end{tabular}}
	\label{breakfast_sota}
\end{table}

\begin{figure*}[t]
\centering
\includegraphics[width=0.9\linewidth]{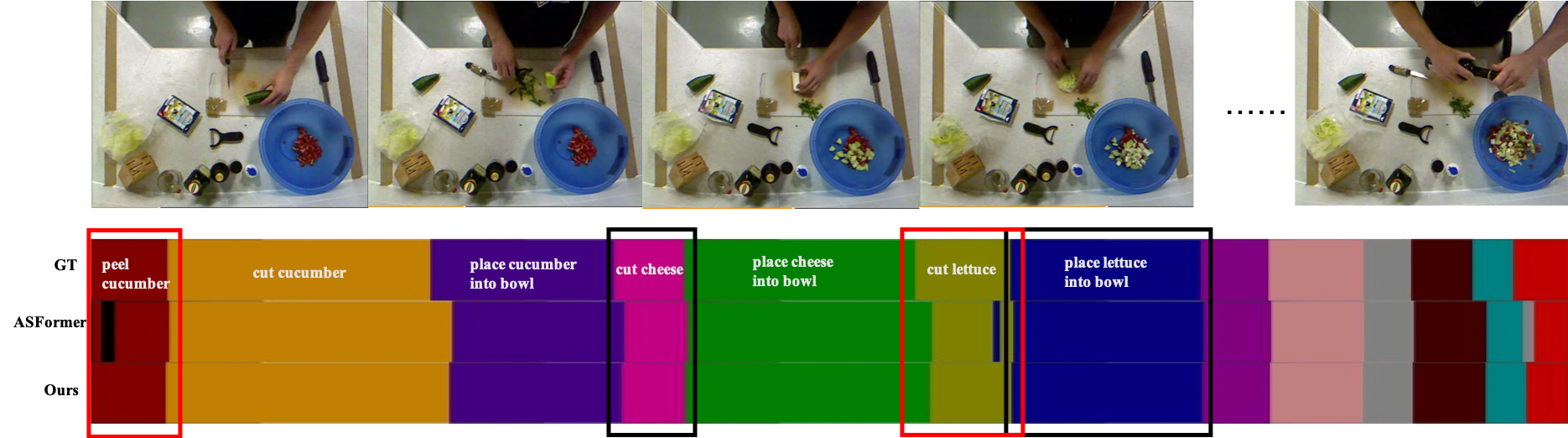}
\vspace{-0.1cm}
\caption{Visualization of the predictions of our model, ASFormer and ground truth (GT).  Only partial segments are shown for clarity. Our segment transformer helps refine both segment labels (red boxes) and boundaries (black boxes).}
\vspace{-0.4cm}
\label{fig:quality}
\end{figure*}



\subsection{Ablation Study}


In this subsection, we perform experiments to develop a detailed understanding of the effectiveness of various components of our proposed architecture on the 50Salads dataset.

\subsubsection{\textbf{Model components}}
Recall that our model consists of two main components, a frame encoder for aggregating frame-level contextual information, and a segment decoder for refining segment representations. To validate the effectiveness of each module, we performed an ablation study which only added the segment decoder to the ASFormer backbone, and aggregated frame-level features directly from the backbone outputs.  As shown in the second row of Table \ref{ablation_components}, performance is significantly improved compared using the backbone alone, demonstrating the segment decoder's utility in refining segment classification and segment boundaries.  Another ablation  is to add the frame encoder, i.e. to compare the performance in the second and third rows of Table \ref{ablation_components}.  The improvement here shows that the frame encoder can aggregate the frame-level contextual information efficiently and provide more useful frame-level features to refine segment representations.

\begin{table}[htb]
  \caption{Ablation on decoder modules on 50Salads dataset.}
  \vspace{-0.1cm}
  \centering
  \resizebox{0.5\textwidth}{!}{
    \begin{tabular}{c|c|c|l|c|c}
   \hline
   Backbone & Segment dec. & Frame enc. & F1@\{10, 25, 50\} & Edit & Acc.\\
   \hline
   \checkmark & - & - & 85.1\; 83.4\; 76.0 & 79.6 & 85.6 \\
   \checkmark & \checkmark & - &  86.5\; 85.1\; 77.6 &  81.3 & 85.8 \\
   \checkmark & \checkmark & \checkmark & \textbf{87.1\; 85.9\; 78.3} &  \textbf{82.4} & \textbf{86.1} \\
   \hline
 \end{tabular}}
 \vspace{-0.3cm}
 \label{ablation_components}
\end{table}


\subsubsection{\textbf{Different segment encoder strategies}} The segment encoder used both frame and category embeddings.  We conducted an ablation on the removal of either type of embedding.  As shown in Table \ref{ablation_query_generation}, dropping the category embedding results in slightly decreased accuracy.  However, removing the frame embedding significantly reduces performance for all metrics.

\begin{table}[htb] 
  \centering
  \caption{Ablation on segment encoder strategies on 50Salads dataset.}
\label{ablation_query_generation}
  \vspace{-0.1cm}
   \resizebox{0.45\textwidth}{!}{
    \begin{tabular}{c|c|l|c|c}
 \hline
  Feature emb. & Category emb. & F1@\{10, 25, 50\} & Edit & Acc.\\
 \hline
 \checkmark  & \checkmark 
 & \textbf{87.1\; 85.9\; 78.3} &  \textbf{82.4} & \textbf{86.1} \\
 \checkmark & $\times$ &  86.2\; 84.9\; 78.1 &  81.3 & 85.9 \\
 $\times$ & \checkmark &  85.6\; 83.9\; 76.8 &  79.6 & 85.7 \\
 \hline
 \end{tabular}}
 \vspace{-0.2cm}
\end{table}

\subsubsection{\textbf{Different segment-frame attention strategies in segment decoder}} 
The segment decoder used segment-frame attention to refine segment classification, based on the hypothesis that frames near a segment provide useful information about the segment itself.  To test this hypothesis we examined in Table \ref{ablation_cross_attention} removing segment-frame attention (row 1), or performing this attention between a segment and all frames in the video (row 2).  The results show that segment-frame attention is useful for improving accuracy, but that attending to nearby frames is better than attending to all frames.  

\vspace{-0.2cm}
\begin{table}[htb] 
  \centering
  \caption{Ablation on segment-frame attention strategies in segment decoder on 50Salads dataset.}
  \label{ablation_cross_attention}
    \vspace{-0.1cm}
   \resizebox{0.45\textwidth}{!}{
    \begin{tabular}{c|l|c|c}
   \hline
   Method  & F1@\{10, 25, 50\} & Edit & Acc.\\
   \hline
   no segment-frame (SF) attention &  85.9\; 84.3\; 77.2 &  80.9 & 85.8 \\
   global SF attention  &  86.6\; 84.8\; 77.7 &  81.5 & 86.0 \\
   local SF attention & \textbf{87.1\; 85.9\; 78.3} &  \textbf{82.4} & \textbf{86.1} \\
   \hline
 \end{tabular}}
 \vspace{-0.2cm}
 \label{ablation_sd}
\end{table}

\subsubsection{\textbf{Different number of layers in segment decoder}}

We performed an ablation study on the optimal number of layers to use in the segment decoder.  As the results in Table \ref{ablation_segment_layer} show, the best performance is obtained with 2 layers.  This shows that using multiple layers is useful, but deep layering is not necessary and its correspondingly high computational cost can be avoided.

\begin{table}[htb] 
  \centering
  \caption{Ablation on number of layers in segment decoder.}
   \resizebox{0.4\textwidth}{!}{
    \begin{tabular}{c|l|c|c}
 \hline
 Layer number  & F1@\{10, 25, 50\} & Edit & Acc.\\
 \hline
     1 &  85.3\; 83.8\; 76.7 &  80.0 & 85.8 \\
 2 & \textbf{87.1\; 85.9\; 78.3} &  \textbf{82.4} & \textbf{86.1} \\
     3 &  86.9\; 85.3\; 78.2 &  81.2 & 85.7 \\
 \hline
 \end{tabular}}
 \label{ablation_segment_layer}
\end{table}

\subsubsection{\textbf{The window size of local cross attention}}

We only performed segment-frame attention on frames which were near each segment, based on the hypothesis that faraway interactions are weak and can be ignored.  To test this assumption, we tested using different window sizes on which to perform attention.  That is, for a window size of $k$, we performed attention between a segment $\sigma$ and all frames within $k$ segments before or after $\sigma$.  As the results in Table \ref{ablation_window_size} show, attention using window size 1 produced the best results, indicating that modeling faraway interactions is not only unnecessary but possibly harmful.

\begin{table}[htb] 
  \centering
  \caption{Ablation on different window sizes for segment-frame attention in segment decoder.}
\label{ablation_sd_window}
   \resizebox{0.35\textwidth}{!}{
    \begin{tabular}{c|l|c|c}
 \hline
 Windows size  & F1@\{10, 25, 50\} & Edit & Acc.\\
 \hline
 1 &  \textbf{87.1\; 85.9\; 78.3} &  \textbf{82.4} & \textbf{86.1} \\
 2 &  86.6\; 85.3\; 78.0 &  81.3 & 85.9 \\
 3 &  86.4\; 84.9\; 78.2 &  81.2 & 85.9 \\
 \hline
 \end{tabular}}
 \label{ablation_window_size}
\end{table}


\subsection{Qualitative Analysis}
Finally, we show in Figure \ref{fig:quality} a qualitative comparison of our segmentation results and the output from ASFormer and ground truth, using a video from the 50Salads dataset.  For clarity, we show only segments from only a portion of the video. We observe from the visualization that our model is able to refine both segment labels and boundaries, and does not produce jittery or unnaturally short segments like ASFormer sometimes does.

\section{Conclusion}
In this paper, we proposed an attention-based action segmentation model called temporal segment transformer, which performs joint segment relations modeling and feature denoising. 
Our method uses a segment-frame attention mechanism to denoise segment representation, and also segment-segment attention to capture temporal dependencies among segments. Our method outperforms all prior approaches on several widely used benchmarks. 

\bibliographystyle{named}
\bibliography{ijcai23}

\end{document}


\maketitle

\section{Pseudo-code description of Our Algorithm}

In order to better understand our algorithm, the pseudo-code is presented below.

\begin{algorithm}[h]
 \SetKwInOut{Input}{Input}\SetKwInOut{Output}{Output}
 \caption{Temporal Segment Transformer}
 \label{alg:cr}
 \SetAlgoLined
 \Input{RGB video features $\mathbf{V}$
      }
      compute frame-level probability distributions $Y$ and frame representations $\Gamma$ using backbone \\
      \For{epoch = 1 to number of epochs}{
        compute segment representations $ \xi$ using segment encoder \\
        compute frame representations $\hat{\Gamma}$ using frame encoder \\
        \STATE \hspace{0.2cm}$ $ \textbf{ for } $i = 1,2,\ldots, L$ \textbf{do} \\
     
        \STATE \hspace{1.0cm} compute $\Xi'_i$ from $\Xi''_{i-1}$ using segment- frame attention \\
        \STATE \hspace{1.0cm} compute $\Xi''_i$ from $\Xi'_i$ using inter-segment attention \\
      }
      compute loss $\mathcal{L}$ using segment classification loss $L_{ce}$ and segment boundary regression $L_{reg}$ \\
      optimize $\mathcal{L}$ using gradient descent \\
 \Output{\\
  segment probability distributions $\tilde{\mathcal{P}}^{mask}$ using segment classification \\
  refined segment boundary ${\mathcal{M}}'$ using offset mapping \\
  frame level predictions $\mathbf{\mathcal{P}}$ using $\tilde{\mathcal{P}}^{mask}$ and ${\mathcal{M}}'$ using mask voting}
\end{algorithm}

\section{More Ablation Study}

In order to develop more detailed understanding of our proposed \textit{temporal segment transformer}, we perform more experiments on 50Salads dataset.

\subsection{Different number of layers in segment decoder}

We performed an ablation study on the optimal number of layers to use in the segment decoder.  As the results in Table \ref{ablation_segment_layer} show, the best performance is obtained with 2 layers.  This shows that using multiple layers is useful, but deep layering is not necessary and its correspondingly high computational cost can be avoided.

\begin{table}[htb] 
  \centering
  \caption{Ablation on number of layers in segment decoder.}
   \resizebox{0.4\textwidth}{!}{
    \begin{tabular}{c|l|c|c}
 \hline
 Layer number  & F1@\{10, 25, 50\} & Edit & Acc.\\
 \hline
     1 &  85.3\; 83.8\; 76.7 &  80.0 & 85.8 \\
 2 & \textbf{87.1\; 85.9\; 78.3} &  \textbf{82.4} & \textbf{86.1} \\
     3 &  86.9\; 85.3\; 78.2 &  81.2 & 85.7 \\
 \hline
 \end{tabular}}
 \label{ablation_segment_layer}
\end{table}

\subsection{The window size of local cross attention}

We only performed segment-frame attention on frames which were near each segment, based on the hypothesis that faraway interactions are weak and can be ignored.  To test this assumption, we tested using different window sizes on which to perform attention.  That is, for a window size of $k$, we performed attention between a segment $\sigma$ and all frames within $k$ segments before or after $\sigma$.  As the results in Table \ref{ablation_window_size} show, attention using window size 1 produced the best results, indicating that modeling faraway interactions is not only unnecessary but possibly harmful.

\begin{table}[htb] 
  \centering
  \caption{Ablation on different window sizes for segment-frame attention in segment decoder.}
\label{ablation_sd_window}
   \resizebox{0.35\textwidth}{!}{
    \begin{tabular}{c|l|c|c}
 \hline
 Windows size  & F1@\{10, 25, 50\} & Edit & Acc.\\
 \hline
 1 &  \textbf{87.1\; 85.9\; 78.3} &  \textbf{82.4} & \textbf{86.1} \\
 2 &  86.6\; 85.3\; 78.0 &  81.3 & 85.9 \\
 3 &  86.4\; 84.9\; 78.2 &  81.2 & 85.9 \\
 \hline
 \end{tabular}}
 \label{ablation_window_size}
\end{table}

\section{More Qualitative results}

In this work, we propose temporal segment transformer for joint segment relation modeling and feature denoising. It exploits a cross attention mechanism on frame representations to denoise segment representation. It adopt self-attention and local-cross attention blocks which are alternatively stacked in segment decoder. In the end, each segment representation is utilized to estimate a refined action label and segment boundary which is further converted to binary segment masks. We generate final segmentation via the voting of the segment masks.

\begin{figure*}[htb]
  \centering
  \includegraphics[width=\linewidth]{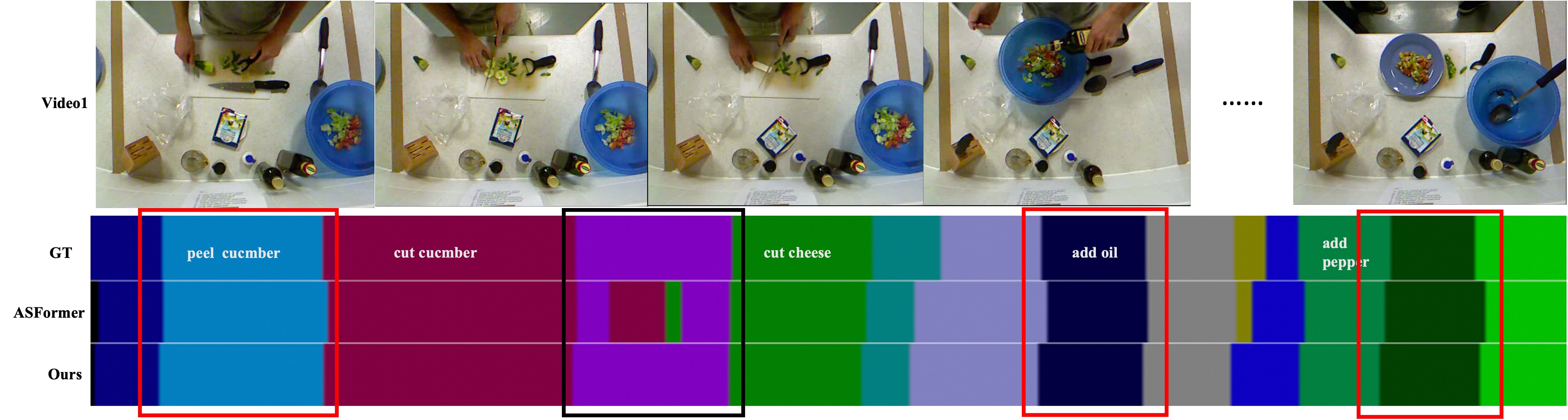}
  \includegraphics[width=\linewidth]{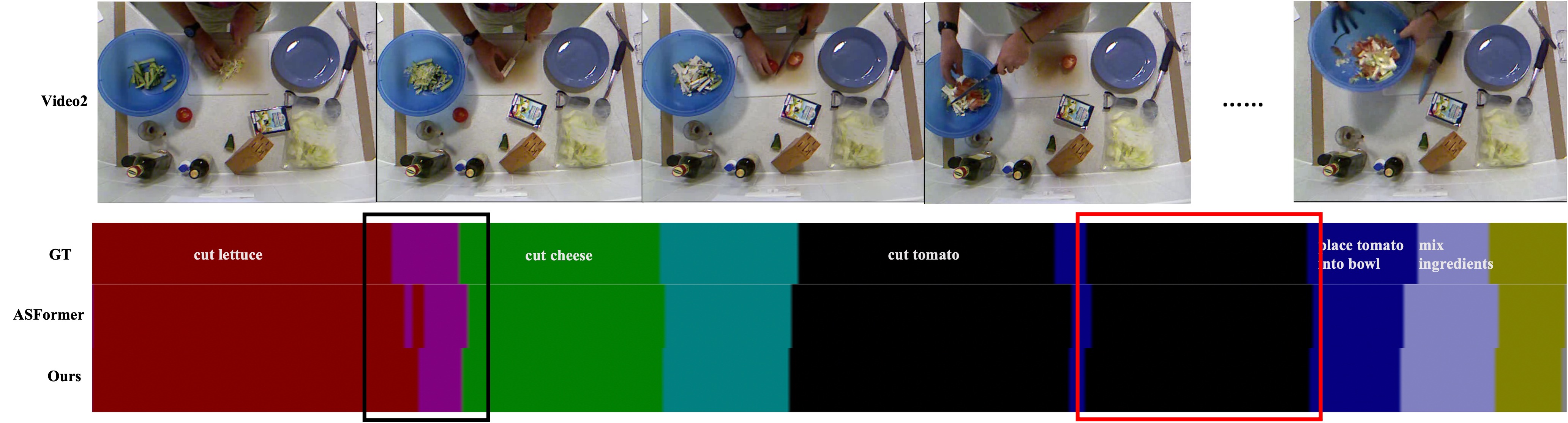}
  \caption{Visualization of two video samples on 50Salad dataset. We compare the predictions of ours, ASFormer and GT, where only partial segments are shown for clarity. We highlight the major improvements with black and red boxes.}
  \label{fig:50Salad}
\end{figure*}

\paragraph{\textbf{50Salads}} As is shown in Figure \ref{fig:50Salad}, we compare the groundtruth, asformer and ours. The black boxes show that our model can refine the boundary base on asformer's results caused by our mask voting strategy, but the degree of boundary improvement is not great enough. The weak sensitiveness probably caused by the lacking of boundary, the current refined boundaries attents more quality of the whole segment. The red boxes demonstrate that the segment classes are refined significantly. The each data dependent segment representation initialized from asformer result are update via local cross attention(enhancing segment representation) and self-attention(considering global context), the final segment embedding is powerful to refine the segment classes.

\begin{figure*}[htb]
  \centering
  \includegraphics[width=1.\linewidth]{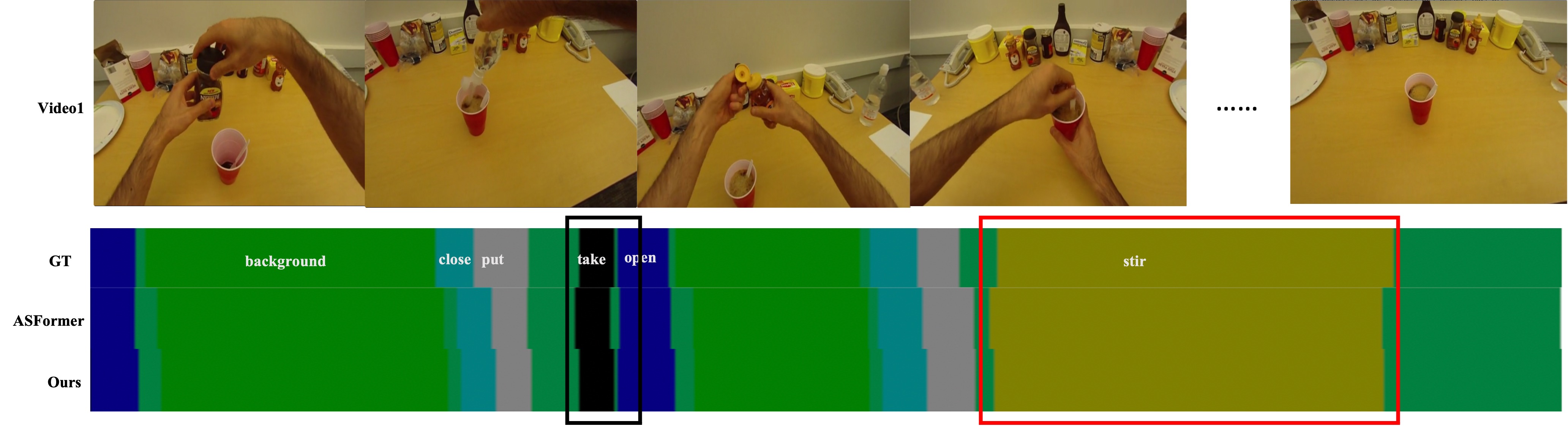}
  \includegraphics[width=1.\linewidth]{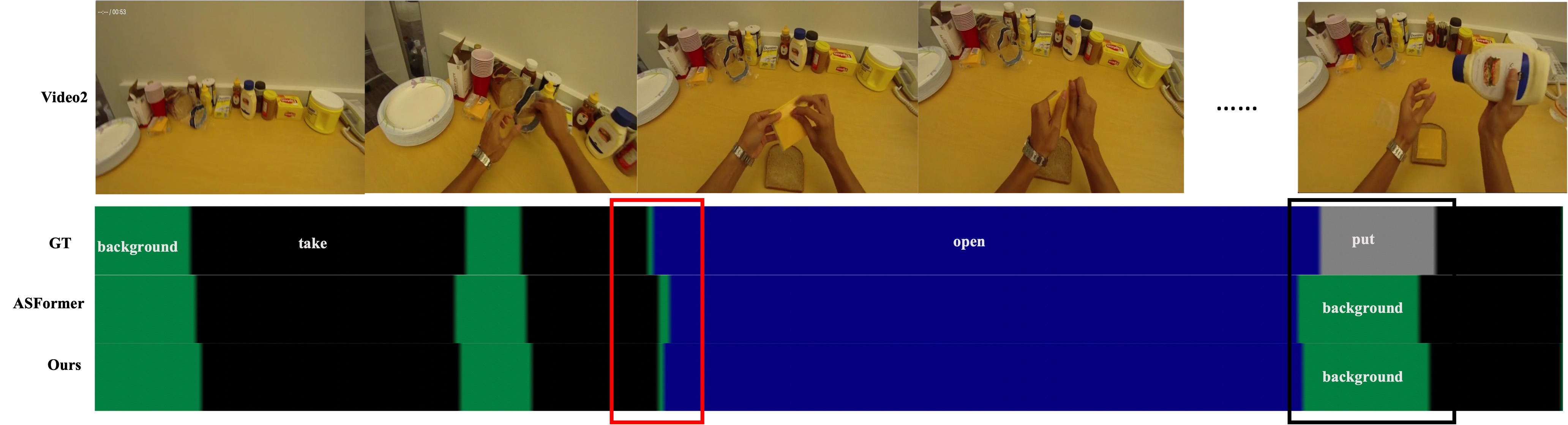}
  \caption{Visualization of two video samples on GTEA dataset. We compare the predictions of ours, ASFormer and GT, where only partial segments are shown for clarity. We highlight the major improvements with red boxes.}
  \label{fig:GTEA}
\end{figure*}

\paragraph{\textbf{GTEA}} As is shown in Figure \ref{fig:GTEA}, The first sample is not suffering from the wrong segment classes, but in our method, we can also refine the segment boundary. It may be difficult for the segment class supervision without  segment boundary. The second sample also refine the segment boundary but fail with segment class refinement in the last video clips. It seems that the self-attention doesn't model the proper global context. 

\begin{figure*}[htb]
  \centering
  \includegraphics[width=1.\linewidth]{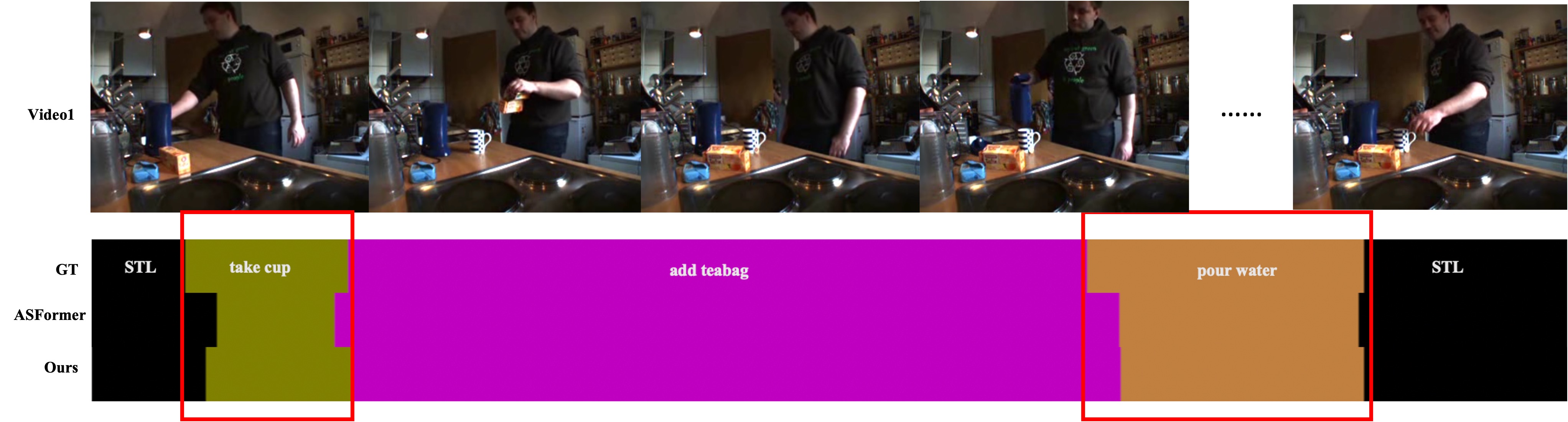}
  \includegraphics[width=1.\linewidth]{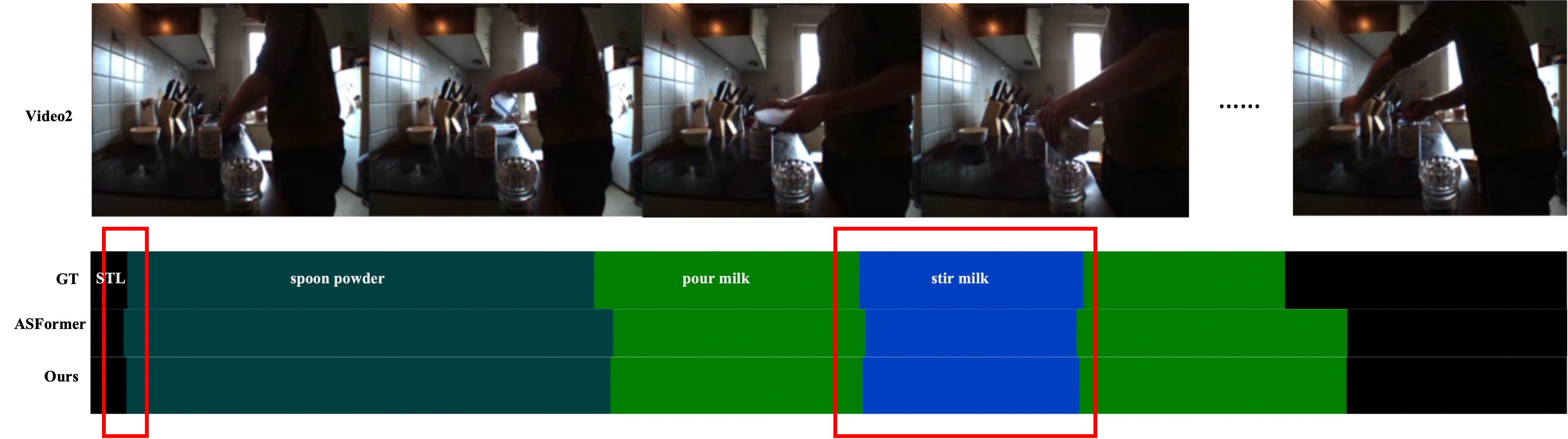}
  \caption{Visualization of two video samples on Breakfast dataset. We compare the predictions of ours, ASFormer and GT, where only partial segments are shown for clarity. We highlight the major improvements with red boxes.}
  \label{fig:breakfast}
\end{figure*}

\paragraph{\textbf{Breakfast}} As is shown in Figure \ref{fig:breakfast}, these two samples are not suffering from wrong segment classes due to the fewer segments, the relationships between each segment can be modeled easily, but it also suffers from the wrong segment boundary. Especially, in the first sample, we refine the take cup significantly on both boundary and refine the right boundary on pour water. That's one of the strengths of our model. But in other segments, the degree of refinement boundary is not enough due to we don't consider the boundary supervision directly or the mask voting strategy is not sensitive to boundary in some cases.

From these visualization results, we can see that our method can refine segment classes significantly and segment boundary in some cases. But our model seems that it's not sensitive to segment boundary significantly, this is our future work for improvement.